\documentclass[conference]{IEEEtran}
\ifCLASSOPTIONcompsoc
  \usepackage[nocompress]{cite}
\else
  \usepackage{cite}
\fi
\usepackage{amsfonts}
\usepackage{wrapfig}
\usepackage[font={small}]{caption}
\ifCLASSINFOpdf
 \usepackage[pdftex]{graphicx}
 \DeclareGraphicsExtensions{.pdf,.jpeg,.png}
\else
   \DeclareGraphicsExtensions{.eps}
\fi
\usepackage{pgfplots}
\usepackage{algorithmic}
\usepackage[fleqn]{amsmath}
\usepackage{amsmath}
\usepackage{tabularx}
\usepackage{graphicx}
\usepackage{multicol, blindtext}
\usepackage{breqn}
\usepackage{multirow}
\usepackage{subfig}
\usepackage{breqn}
\usepackage{color}
\usepackage{graphics}
\usepackage{colortbl}
\usepackage{tabulary}
\usepackage{etoolbox}
\usepackage{colortbl}
\usepackage{bigstrut}
\usepackage{multirow}
\usepackage{array}
\usepackage{tabularx}
\usepackage{booktabs}
\usepackage{hhline}
\usepackage{dsfont}

\usepackage[a4paper]{geometry}
\newgeometry{right=0.62in,left=0.62in,top=0.75in,bottom=1in}

\title{How Secure is Distributed Convolutional Neural Network on IoT Edge Devices?}

\vspace{0mm}

\author{
    \IEEEauthorblockN{
    Hawzhin Mohammed\IEEEauthorrefmark{1},
    Tolulope A. Odetola\IEEEauthorrefmark{1},
    Syed Rafay Hasan\IEEEauthorrefmark{1},
    }

    \vspace{0mm}
    \IEEEauthorblockA{
    \IEEEauthorrefmark{1}Department of Electrical and Computer Engineering, Tennessee Technological University, Cookeville, TN 38505, USA}

\vspace{0mm}
}

\begin{document}
\maketitle
\vspace{0mm}
\begin{abstract}
Convolutional Neural Networks (CNN) has found successful adoption in many applications. The deployment of CNN on resource-constrained edge devices have proved challenging. CNN distributed deployment across different edge devices has been adopted. In this paper, we propose Trojan attacks on CNN deployed across a distributed edge network across different nodes. We propose five stealthy attack scenarios for distributed CNN inference. These attacks are divided into trigger and payload circuitry. These attacks are tested on deep learning models (LeNet, AlexNet). The results show how the degree of vulnerability of individual layers and how critical they are to the final classification.
\end{abstract}


\vspace{0mm}
\section{Introduction}
\label{Introduction}
\begin{figure}[t]
\centering
\includegraphics[scale=0.32]{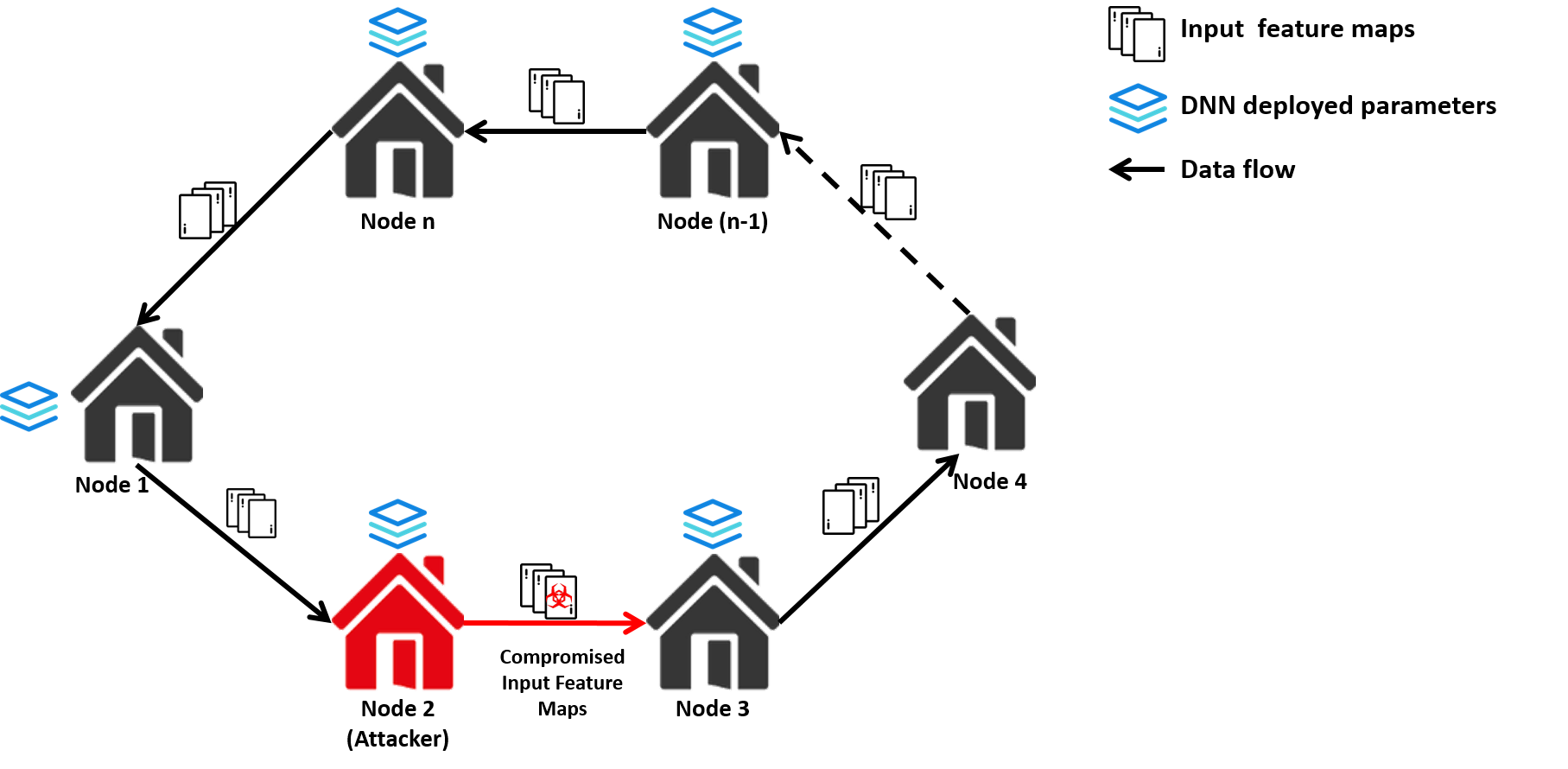}
\caption{Community Area Network}
\vspace{0mm}
\label{fig:CAN}
\end{figure}

\vspace{0mm}
Internet of Things (IoT) edge devices used in conjunction with Convolutional Neural Networks (CNN) is increasingly adopted in many applications such as smart homes, smart cities, autonomous vehicles, and healthcare \cite{zhou2019distributing}.
CNN inference incurs large computation and memory overhead. To combat this overhead, cloud-based CNN inference has been adopted, but it raises security and privacy concerns \cite{odetola20192l} along with an increase in communication latency \cite{zhou2019distributing}.
To avoid this latency, edge computing has been introduced where CNN inference can be divided between resource-constrained node devices and edge servers \cite{li2018edge}. For some of the applications like smart homes and community area networks (CAN) (see Fig. \ref{fig:CAN}) there is a need of exploring edge computing only using resource-constrained node level edge devices (ND). We have seen in the past that researchers have used mobile crowd computing (MCC) \cite{guo2015mobile} to collaborate for a task using a cluster of mobile devices. Along the same lines recently, the idea of using multiple resource constrained devices to implement distributed CNN has been explored \cite{mao2017modnn, mao2017mednn}. The deployment of CNN on a locally distributed edge network offers the advantage of higher privacy and lower dependency on network bandwidth \cite{mao2017modnn}.

\subsubsection{What is the Security Vulnerability in Distributed CNN}
Some of the attacks against CNN hardware accelerators discussed in literature includes: Liu $et. al$ in \cite{liu2017trojaning}  proposes a Trojan attack on neural networks that generate samples of the input from the weights of the pre-trained CNN model to form a trigger circuit for a payload that induce malicious behavior of the neurons of the CNN. Clements $et. al$ \cite{clements2018hardware} introduces a framework carried CNN hardware accelerators where a specific input image trigger perturbations that are added to targeted layers CNN to cause mis-classification in the final layer. These above approaches assumes the attacker has knowledge of the type of input image, full access to all the weights of the CNN model, full information of the CNN architecture and can see the direct consequence of the hardware attacks. This is not the case with the distributed deployment and execution of CNN models across multiple devices. Hence Distributed CNN is traditionally considered more secure. One of the recent attacks \cite{hailesellasie2019vaws} shows that attacks on parameters of only one of the CNN layers can lead to accuracy loss. Although their work requires resource-intensive brute-force approach but it has a direct implication to the security of distributed CNN on ND. With this intuition, we analyzed LeNet \cite{lecun1998gradient} to see the effect of replacing the weights of the first convolutional layer with a very small number. This result showed us that such a replacement can cause a drastic loss in accuracy. Hence, this observation warrants the following research question to be answered \textit{Can an attacker utilize only the information that one node has with respect to CNN to launch a stealthy attack}.

\subsubsection{What are the Research Challenges}
In this work, we argue that such distributed designs are vulnerable to security attacks as the case of distributing CNN to unknown parties may lead to a compromised malicious node. The attacker has the hold of ND only and in this distributed CNN the ND only has access to weights associated with that layer, input features of the layer and the output that it generates. Hence, it is challenging to associate how to make use of only this information to generate meaningful and stealthy attacks.

\begin{figure*}[t]
\vspace{0mm}
\centering
\includegraphics[scale=0.6]{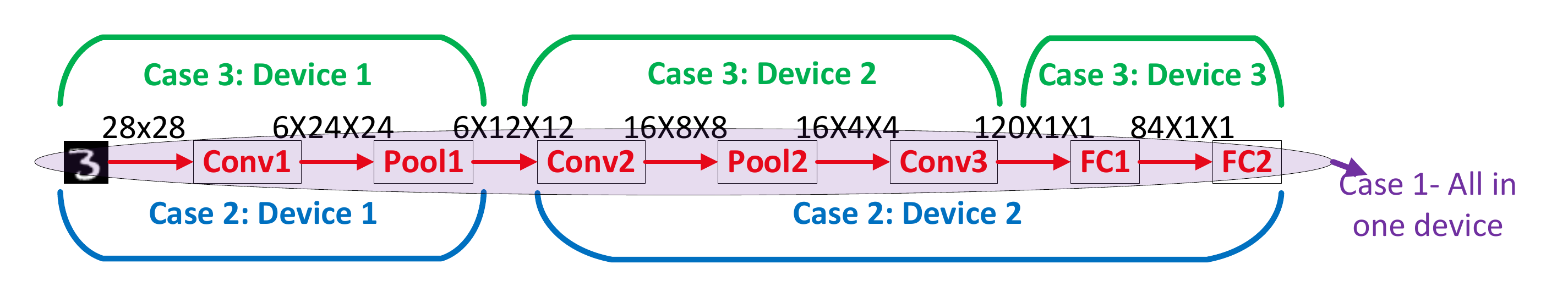}
\caption{CNN Layer Partitoning}
\vspace{0mm}
\label{fig:Partiton}
\end{figure*}

\subsubsection{Novel Contributions}
In this work our novel contributions are as follows:
\begin{itemize}
\item A layer by layer analysis to identify the most vulnerable layer of the distributed CNN located in untrusted ND.
\item We came up with a series of low overhead intermittent attacks that can potentially lead to mis-classification.
 	\end{itemize}

The remainder of this paper is organized as follows: Section \ref{CNN Layer-to-Node Partitioning} provides some preliminary information on distributed CNN, partitioning of the layers. Section \ref{Network and Threat Model} discusses the network and threat model. Section \ref{Attack Scenarios} describes the proposed attack scenarios. Section \ref{Proposed Methodology} shows how to implement attack scenarios stealthily. Section \ref{Experimental Setup} discusses the experimental setup. Section \ref{Results and Discussions} provides the results and discussions. Section \ref{Comparison With State-of-the-Art} discusses the comparison with the state-of-the-art. Literature review is provided in Section \ref{Literature Review}. Section \ref{Conclusion} concludes the paper.

\vspace{0mm}
\section{Distributed CNN: Partitioning of the Layers}
\label{CNN Layer-to-Node Partitioning}
\vspace{0mm}

CNN architectures consist of convolutional, pooling, normalization, non-linear activation, and fully connected layers. Convolutional layers are the most computation-intensive while the fully connected layers are more memory intensive compared to other layers \cite{mao2017modnn}. To achieve efficient distributed CNN inference on multiple resources constrained nodes, CNN has to be partitioned based on the capabilities of the nodes involved in the network.
Several partitioning schemes have been adopted in literature for optimum partitioning based on the computation, memory, and communication (bandwidth) capabilities of the nodes in the network \cite{mao2017modnn, mao2017mednn, zhao2018deepthings, zhou2019distributing}. As is the case with any implementation, our distributed CNN has some unique features, though the overall idea is inspired by the above-mentioned references.
Fig. \ref{fig:Partiton} shows an example case of our overall approach for Layer-based partitioning. In Fig. \ref{fig:Partiton} a CNN model (LeNet) is considered, Conv1, Conv2, Conv3 represents three convolutions layers and FC1, FC2 shows two fully connected layers. In case 1, it is shown that all the seven layers of CNN models are undivided. Case 2 shows that since Conv1, and Pool1 require the most computation hence the partitioning is done such that they are allocated to one device and the rest of the network is placed on the Device 2. Finally, in the third case, we further sub-partitioned the layers in Device 2 into two devices.
We scaled our technique to other layers based on the computation and memory requirement along with the total number of output features each layer needs to transfer. Since this is not the core of our work hence the details of partitioning are not further discussed in this paper.

\begin{figure*}[t]
\vspace{0mm}
\centering
\includegraphics[scale=0.4]{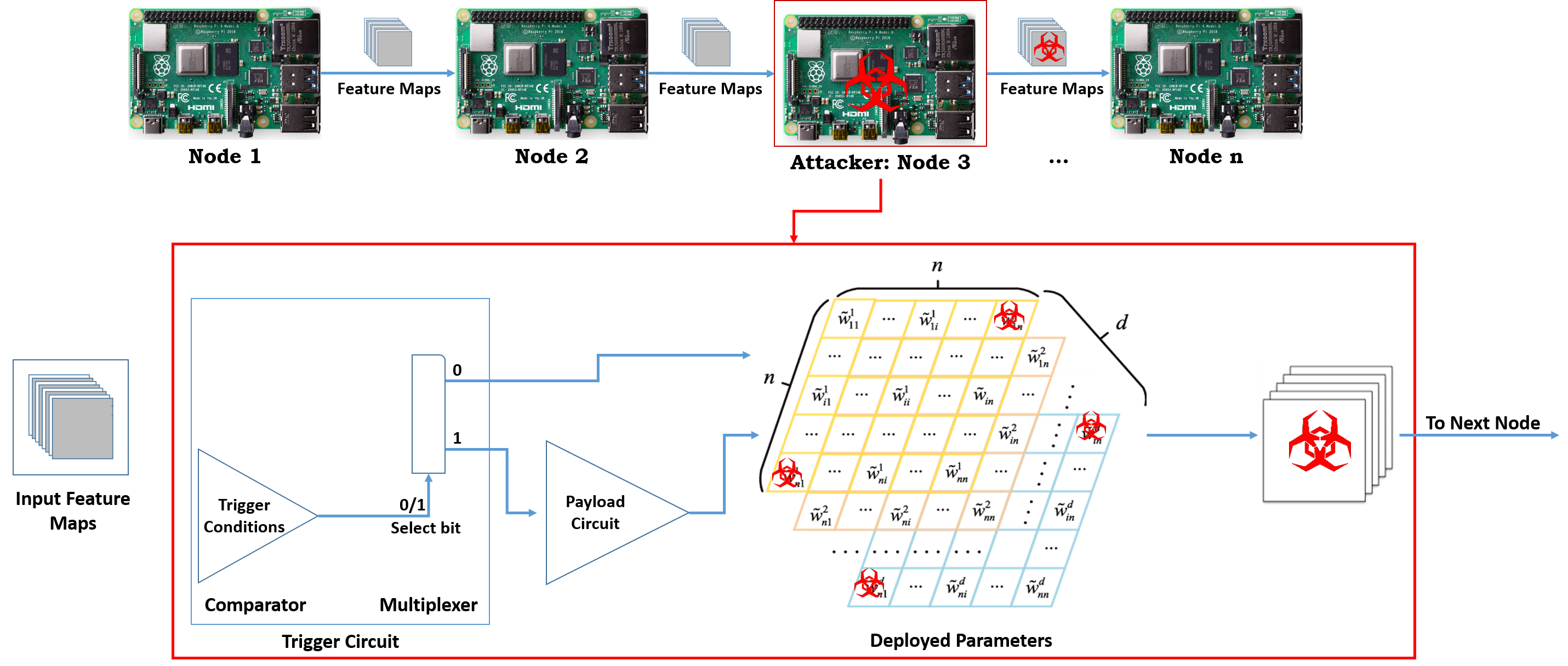}
\caption{Proposed Attack Scenarios}
\vspace{0mm}
\label{fig:attacker}
\end{figure*}
\vspace{0mm}
\section{Network and Threat Model}
\label{Network and Threat Model}
\vspace{0mm}

\vspace{0mm}
\subsection{Network Model}
\label{Network Model}
\vspace{0mm}
The network model consists of IoT devices connected together forming a Wireless Local Area Network (WLAN).
Each IoT node device connected to its neighbor through wireless communication technology \cite{mao2017modnn}.
Conceptual network topology for CAN where the IoT node devices are connected together is shown in Fig. \ref{fig:CAN}.
Multiple IoT devices collaborate to form the WLAN (inspired from crowd-computing) of a locally distributed edge IoT-based network containing layers of a CNN to perform classification. To perform classification on the locally distributed network, the CNN layers are distributed across different IoT node level edge devices (NDs) in the WLAN. These IoT node devices may belong to different vendors and may have different hardware \& software characteristics.

\vspace{0mm}
\subsection{Threat Model}
\label{Threat Model}
\vspace{0mm}
In the threat model we assume these NDs may belong to an attacker with malicious intentions. The attacker has no knowledge of the full architecture of the CNN, input image to the CNN and final classification. The attacker only has access to the parameters deployed on a particular ND, and the input features passed from previous ND for node-wise distributed operation (to do part of the computation for distributed CNN) to generate the output of the node-wise operation.
Node-wise operation may be convolution, pooling, normalization, activation and/or classification depending on the parameters deployed on the node.

\vspace{0mm}
\section{Proposed Attack Scenarios}
\label{Attack Scenarios}
\vspace{0mm}
In this section, the attacks that have been performed on the CNN layer at the attacker node have been explained. As elaborated earlier, after partitioning of the layers for distributed CNN inference, the attacker node has access to the parameters associated with its ND explained in Section \ref{Threat Model}.

\vspace{0mm}
\subsection{A Novice Scalar Attack}
\label{Scalar Attack}
\vspace{0mm}
Since all the weights (parameters) follow Gaussian distribution, therefore, hence multiplying each parameter with an arbitrary scalar value will lead to a shift in the mean value of the Gaussian Distribution to cause mis-classification. Bearing this in mind in this attack, a scalar multiplication of all the weights is applied, where $scalar$ $\in$ \{0.6, 0.8, 1.2, 1.4\} $\forall$ $x_i \in X$ and $X$ is values of the parameters. The attacker's intention is to reduce or increase the parameters by a certain percentage. In this work four percentage values have been chosen, two to increase the value of the parameter and two to decrease. Increment and decrement in values are made up to 20\% and 40\%.

\vspace{0mm}
\subsection{Random Scale Attack}
\label{Random Attack}
\vspace{0mm}
In this attack, randomly chosen values are multiplied by the values of the parameter. The random values that have been chosen by the attacker are Gaussian distributed between 0 and 1.
The rationale behind this attack is since it is computationally and software-wise less intensive to introduce such attack even on the fly. Further discussion is made on the overhead of such an attack in the Section \ref{Results and Discussions}.

\vspace{0mm}
\subsection{Polarity Switch Attack}
\label{Sign Flip Attack}
\vspace{0mm}
In the polarity switch attack, the attacker flips all the signs of the values of the parameters. Any value with a positive sign has been switched to negative and all negative signs have been switched to positive.
In this attack, the attacker does not change the values of the parameters, where $flip$ $=$ \{-1\} $\forall$ $x_i \in X$ and $X$ is values of the parameters.

\vspace{0mm}
\subsection{Maximum Minimum Swap Attack}
\label{Parameter Swapping Attack}
\vspace{0mm}
By maximum-minimum swap attack, we mean that the attacker has changed the index of the maximum and minimum values of the parameters. The attacker does not change the statistical properties of the parameters' matrix (i.e. the normal distribution, the mean, the variance, stayed unchanged). Since there is no heavy computation involved therefore this attack is less on overhead and at the same time affects the accuracy.

\vspace{0mm}
\subsection{Parameter Statistical Attack}
\label{Parameter Statistical Attack}
\vspace{0mm}
The values of the parameters have a normal distribution of $X \sim \mathcal{N}(\mu,\,\sigma^{2})\,$, where $X$ is values of the parameters.In this attack, all the values of the parameters $X$ have been changed by the attacker with other new values $X'$ that have the same normal distribution property $\mathcal{N}(\mu,\,\sigma^{2})\,$. The new values have the same mean $\mu$ and the same standard deviation $\sigma^{2}$ like the original values. This attack is heavy on memory but if a defender is monitoring the parameters based on their statistical properties then it is a very stealthy attack.

\vspace{0mm}
\section{How to Implement Attack Scenarios Stealthily}
\label{Proposed Methodology}
\vspace{0mm}
The proposed methodology is shown in Fig. \ref{fig:attacker} depicts the partitioning of a CNN on IoT edge devices in a CAN. In Fig. \ref{fig:attacker}, the input images are collected and passed from Node $1$ where the first layer is situated to Node $n$ where the final classification layer is located. In the methodology, it is assumed that Node $1$ and Node $n$ are trusted nodes and intermediary nodes (Node $2$ to Node $n-1$) are untrusted nodes. In Fig. \ref{fig:attacker}, Node $3$ is assumed to be an attacker with the malicious intent of causing mis-classification at the final layer in Node $n$.

In Node $3$, the attacker has access to the incoming input feature maps which is the output of the prior node (Node $2$). The attacker has access to the trainable parameters (weights and biases) deployed within the Node. The attack also has access to the output feature maps at Node $3$ which serves as input to latter nodes (Node $4$).

The attacks perpetrated in this work are divided into trigger and payload circuits. The trigger can be used to activate different payload circuits to implement the attack scenarios. The trigger can be a software-based stimulus or hardware generated signal.

\subsection{Trigger Design}
When the distributed CNN is established, it is presumed that the information required for each ND is also provided. This includes the information for the deployed layer's parameters. Based on this information, a malicious software code in the processor can establish its triggering condition. Since the attacker, ND has complete access to the given information and the device hence it is feasible for the attacker to make this trigger an intermittent one to make it more stealthy.

\subsection {Payload Design}
The payload circuit is activated by the trigger with the goal of compromising the output feature maps at the node that serves as input to the next node on the line of the network with the objective of causing mis-classification at the final layer as stealthily as possible. In this work, five payload mechanisms are discussed to achieve the attacks discussed in Section \ref{Attack Scenarios}.

For all the attacks where we need to multiply the parameters with either a scalar number or random number, our methodology used the following method. In order to reduce the timing overhead, we implemented the attack using an additional buffer. Once triggered this additional buffer sends out the corrupted parameters instantly. Hence, we achieve reduced timing overhead as shown in Section \ref{Results and Discussions}. In the attack where we required swapping of the values, the maximum and minimum values of the attack are swapped in place with the use of an intermediate buffer. This attack only uses a comparator and does not require heavy usage of multipliers.

\vspace{0mm}
\section{Experimental Setup}
\label{Experimental Setup}
\vspace{0mm}
In this section, we explain the experimental setup that has been conducted to confirm the effectiveness of the proposed methodology.

\vspace{0mm}
\subsection{Device Type(s)}
\label{Device Type}
\vspace{0mm}
The experimental setup consists of Raspberry Pi 3 Model B. Its specification is: 1) a CPU with a speed of 1.2GHz, 2) a RAM of size 1GB, 3) a wireless LAN interface and a Bluetooth Low Energy (BLE) interface on the board, and 4) a fast Ethernet interface (100 base-Ethernet) \cite{aldrich2017raspberry}.
These devices have been used because they are one of the most commonly used platforms in modern IoT-based network experiments \cite{hamdan2019iot, pavithra2015iot, vujovic2015raspberry}.

\vspace{0mm}
\subsection{Communication setup}
\label{Communication setup}
\vspace{0mm}
Each node connected to the other node using the Ethernet interface (this can also be translated into the WiFi interface as well). Each node has a unique  IP address and a unique port number for communication purposes. After processing the data, each node sends the result, i.e. the blobs of the final layer to the next node through the communication interface.

\vspace{0mm}
\subsection{Experimental Scenarios}
\label{Experimental Scenarios}
\vspace{0mm}
In the experiment, two CNN networks have been used to show the effects of the attacks, namely, LeNet CNN \cite{lecun1998gradient} with five layers (three convolutions and two fully connected) that has been used for digit handwritten recognition, and AlexNet \cite{krizhevsky2012imagenet} with eight layers (five convolutions and three fully connected) for imageNet dataset \cite{deng2009imagenet} recognition. The CNN has been partitioned and divided among the NDs, each ND gets a part of the CNN that depends on its capability. The attacker node has access to a subset of parameters i.e. to the parameters of the layer that it receives. Then the attacker starts the attack by manipulating the weights of the received layer. All the attacks mentioned in Section \ref{Attack Scenarios} have been implemented. Depending on the type of attack, that attacker changes the weights of the received layer, as shown in Fig. \ref{fig:attacker}.
Finally, the attacker receives the data blobs (output of previous IoT ND), processes the data, and then sends the new blobs (output feature) to the next IoT ND.

\vspace{0mm}
\section{Results and Discussions}
\label{Results and Discussions}
\begin{figure}[t]
\setlength{\abovecaptionskip}{0mm}   
\setlength{\belowcaptionskip}{0mm}   
\centering
\includegraphics[scale=0.89]{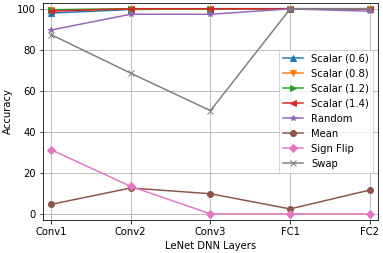}
\caption{Comparison among all attacks on LeNet CNN}
\vspace{0mm}
\label{fig:lenet}
\end{figure}

\begin{figure}[t]
\setlength{\abovecaptionskip}{0mm}   
\setlength{\belowcaptionskip}{0mm}   
\centering
\includegraphics[scale=0.89]{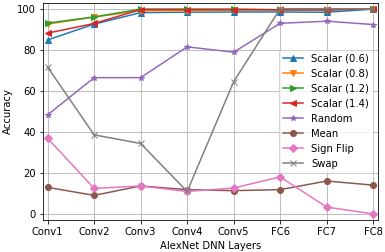}
\caption{Comparison among all attacks on AlexNet CNN}
\vspace{0mm}
\label{fig:alexnet}
\end{figure}

\vspace{0mm}
Five different attacks on the values of the parameters of a CNN layer have been proposed in this work.
Each layer of the CNN has parameters that are trained to perform high accuracy classification on the input data.
These layers can be convolution layer or fully connected layer. Convolution layers usually require high computation operation and a fully connected layer requires high memory resources \cite{abtahi2018accelerating}. From Figs. \ref{fig:lenet} and \ref{fig:alexnet}, it can be seen that some attack affects the computation-centric layers, i.e., the convolution layer, other attacks affect memory-centric layer, i.e., fully connected layer, and some other attack affects both types of the layer equally.

The scalar attack results on LeNet CNN and AlexNet CNN are shown in Fig. \ref{fig:lenet} and Fig. \ref{fig:alexnet}, respectively.
It can be seen from the figures that the scalar attack has more effect on computation-centric layers, i.e., convolution layer. The $scalar$ value $\in$ \{0.6, 0.8, 1.2, 1.4\} has an effect on the accuracy more on the layers that are near the input data. From Figs. \ref{fig:lenet} and \ref{fig:alexnet}, the accuracy is less at convolution 1 and 2 compare to layers near to output of the network, i.e., the fully connected layer 1 and 2 for the scalar attack.
The random attack is shown in Figs. \ref{fig:lenet} and \ref{fig:alexnet} follows the scalar attack pattern but with more effect on the accuracy. The layers near the input data, i.e., computation-centric layers or the convolution layers affect the accuracy more than the layers near the output of the CNN network, i.e., memory-centric layers or the fully connected layers.

On the other hand, the sign flipping attack has more effect on the memory-centric layers, the fully connected layer.
As can be seen from Figs. \ref{fig:lenet} and \ref{fig:alexnet}, the accuracy at the layers near the output of the network is almost zero, but with the same attack on convolution layer 1 still, the network has around 40\% accuracy.
The swap attack, where the attacker swap the index of the maximum and minimum values at the layer, has more effect on the accuracy of the convolution layers.
As each convolution has multiple channels and each channel has it's maximum and minimum values. When the number of channels increases the swap attack effect increases on the accuracy. The fully connected layer has one matrix with one maximum and one minimum value when the attacker swaps them, the accuracy is not going to be affected much.

The parameter statistical attack affects all the layers equally. Figs. \ref{fig:lenet} and \ref{fig:alexnet} shows that the mean attack affects the accuracy of a small network like LeNet and big network like AlexNet is less than 20\% for all the layers.
In Fig. \ref{fig:alexnet}, there is a data outlier for swap attack at the convolution layer 5.
The effect of the attack on this layer is equivalent to convolution layer 1, where convolution layer 5 has more parameter and perform more computational operation compare to layer 1.
Tables \ref{tab:lenet} and \ref{tab:alexnet} summarize all the accuracy across all the layers for LeNet and AlexNet network.

Tables \ref{tab:timelenet} and \ref{tab:timealexnet} summarize the overhead required by the attacker to change the values of the parameters. As the number of parameter increases the time needed by the attacker increases, for example, convolution layer 3 and fully connected layer 6 in AlexNet has higher parameter compared to other layers. Also, for LeNet the convolution layer 3 takes longer time to attack as the number parameters in this layer are higher compared to other layers. The parameter statistical attack needs lesser time, as the attacker needs to overwrite the values of the parameters to the memory only but in other attack scenarios the attacker needs to read from the memory, do the attack operation, then write back to the memory which leads to a longer time.

\begin{table}[h]
  \centering
  \caption{Accuracy across layer by layer for LeNet CNN with the five attacks.}
    \begin{tabular}{|l|c|c|c|c|c|c|c|c|}
    \toprule
    \hspace{-20mm} & \hspace{-2mm} 0.6 \hspace{-2mm} & \hspace{-2mm} 0.8 \hspace{-2mm} & \hspace{-2mm} 1.2 \hspace{-2mm} & \hspace{-2mm} 1.4 \hspace{-2mm} & \hspace{-2mm} Random \hspace{-2mm} & \hspace{-2mm} Mean \hspace{-2mm} & \hspace{-2mm} Flip \hspace{-2mm} & \hspace{-2mm} Swap \hspace{-2mm} \\
    \midrule
    \midrule
    Conv1 & 97.9 & 99.4 & 99.5 & 98.8 & 89.6 &  4.6 & 31.3 & 87.4 \\
    Conv2 & 99.6 & 99.8 & 100  & 100  & 97.3 & 12.6 & 13.4 & 68.6 \\
    Conv3 & 100  & 100  & 100  & 100  & 97.3 &  9.8 &    0 & 50.3 \\
    FC1   & 100  & 100  & 100  & 100  & 99.9 &  2.4 &    0 & 99.9 \\
    FC2   & 100  & 100  & 100  & 100  & 98.8 & 11.6 &    0 & 100  \\
    \midrule
    \end{tabular}%
    \vspace{0mm}
  \label{tab:lenet}%
\end{table}%
\begin{table}[h]
  \centering
  \caption{Accuracy across layer by layer for AlexNet CNN with the five attacks.}
    \begin{tabular}{|l|c|c|c|c|c|c|c|c|}
    \toprule
    \hspace{-20mm} & \hspace{-2mm} 0.6 \hspace{-2mm} & \hspace{-2mm} 0.8 \hspace{-2mm} & \hspace{-2mm} 1.2 \hspace{-2mm} & \hspace{-2mm} 1.4 \hspace{-2mm} & \hspace{-2mm} Random \hspace{-2mm} & \hspace{-2mm} Mean \hspace{-2mm} & \hspace{-2mm} Flip \hspace{-2mm} & \hspace{-2mm} Swap \hspace{-2mm} \\
    \midrule
    \midrule
    Conv1 & 84.8 & 92.5 & 93.0 & 88.1 & 48.3 & 12.9 & 36.8 & 71.5 \\
    Conv2 & 92.5 & 95.9 & 96.0 & 92.8 & 66.4 &  9.0 & 12.4 & 38.5 \\
    Conv3 & 98.1 & 99.1 & 98.8 & 99.6 & 66.4 & 13.6 & 13.6 & 34.3 \\
    Conv4 & 98.3 & 99.1 & 99.8 & 99.6 & 81.4 & 11.8 & 10.9 & 10.9 \\
    Conv5 & 98.4 & 99.3 & 99.8 & 99.6 & 78.8 & 11.3 & 12.6 & 64.4 \\
    FC6   & 98.4 & 99.4 & 99.5 & 99.5 & 92.9 & 11.8 & 18.0 & 100  \\
    FC7   & 98.3 & 99.4 & 99.5 & 99.5 & 93.9 & 16.0 &  3.3 & 100  \\
    FC8   & 99.8 & 100  & 99.9 & 99.9 & 92.3 & 14.0 &    0 & 99.6 \\
    \midrule
    \end{tabular}%
    \vspace{0mm}
  \label{tab:alexnet}%
\end{table}%
\begin{table}[h]
  \centering
  \caption{Time required to perform the attack on LeNet in (ms).}
    \begin{tabular}{|l|c|c|c|c|c|c|c|c|}
    \toprule
    \hspace{0mm} & \hspace{0mm} Scalar  \hspace{0mm} & \hspace{0mm} Random \hspace{0mm} & \hspace{0mm} Mean \hspace{0mm} & \hspace{0mm} Flip \hspace{0mm} & \hspace{0mm} Swap \hspace{0mm} \\
    \midrule
    \midrule
    Conv1 &     5.117 &     5.338 &   2.767 &    5.485 &   2.044 \\
    Conv2 &    80.006 &    84.791 &  30.045 &   87.492 &  27.196 \\
    Conv3 & 1018.723  & 1083.635  & 383.760 & 1105.240 & 541.260 \\
    FC1   &  261.082  &  275.140  &  58.463 &  286.935 &   2.445 \\
    FC2   &   21.811  &   23.398  &   5.774 &   24.540 &   1.745 \\
    \midrule
    \end{tabular}%
    \vspace{0mm}
  \label{tab:timelenet}%
\end{table}%
\begin{table}[h]
  \centering
  \caption{Time required to perform the attack on AlexNet in (sec.).}
    \begin{tabular}{|l|c|c|c|c|c|c|c|c|}
    \toprule
    \hspace{0mm} & \hspace{0mm} Scalar  \hspace{0mm} & \hspace{0mm} Random \hspace{0mm} & \hspace{0mm} Mean \hspace{0mm} & \hspace{0mm} Flip \hspace{0mm} & \hspace{0mm} Swap \hspace{0mm} \\
    \midrule
    \midrule
    Conv1 &  1.284 &  1.327 &  0.509 &  1.360 &  0.096 \\
    Conv2 & 11.425 & 11.978 &  4.646 & 12.211 &  3.935 \\
    Conv3 & 34.689 & 35.410 & 13.871 & 36.929 & 31.262 \\
    Conv4 & 25.943 & 26.719 & 10.195 & 27.263 & 23.611 \\
    Conv5 & 17.403 & 17.708 &  6.951 & 18.280 & 15.717 \\
    FC6   & 980.930&1161.937&215.432 &1106.786&  2.597 \\
    FC7   & 434.290& 513.975& 95.752 & 513.033&  1.189 \\
    FC8   &  0.950 &  1.015 &  0.229 &  1.039 &  0.004 \\
    \midrule
    \end{tabular}%
    \vspace{0mm}
  \label{tab:timealexnet}%
\end{table}%
\vspace{0mm}
\section{Comparison With State-of-the-Art}
\label{Comparison With State-of-the-Art}
\vspace{0mm}

Since distributed CNN is a new concept so no direct work can be found to compare our attack techniques with others. However, some of the existing attacks on deep learning architecture are compared against our proposed attacks. In \cite{hailesellasie2019vaws} Hailesellasie $et. al$ utilize a framework that analyzes the sensitivity of weights of each respective layers and how critical they are to the accuracy of the CNN. The sensitive weights are then attacked during run-time to cause mis-classification. The threat model adopted in this framework gives the attacker access to all weights of all the layers of the CNN model. The threat model did not consider a situation where the attacker has partial access to the weights and has no knowledge of the type of inputs and outputs of the CNN model.
Zou $et.  al$ \cite{zou2018potrojan} proposes PoTrojan which has a trigger and a payload. When the Trojan is activated, the payload adds small neurons and synapses to the targeted layer(s) of the CNN layers. The threat model assumes the attacker has access to all weights of the CNN model. The attacker can see the consequence of the attack. Our proposed work is different since their threat model does not consider parallel execution of CNN models on multiple devices, where the attacker has partial access to the weights and has no knowledge of the type of inputs and outputs of the CNN model.

Clements $et. al$ \cite{clements2019hardware} proposes an attack on neural networks that targets the activation function of a neural network. The threat model used in this attack assumes all weights of the CNN model can be accessed by the attacker. The attacker can see the impact of the attack on mis-classification of the model. The threat model does not take into consideration parallel execution of CNN models on multiple devices, where the attacker has partial access to the weights and has no knowledge of the type of inputs and outputs of the CNN model.

\vspace{0mm}
\section{Literature Review}
\label{Literature Review}
\vspace{0mm}

Zhou $et. al$ \cite{zhou2019distributing} proposes a framework that explores the parallel execution of the CNN inference phase across multiple resources constrained heterogeneous devices. The framework uses a dynamic programming based search algorithm to compute the optimal partition and parallelization of a CNN. Their framework also proposes a CNN acceleration framework that adaptively computes the resources and network conditions among heterogeneous devices.
Zhao $et. al$ \cite{zhao2018deepthings} proposes a framework for adaptively distributing CNN inference among resource-constrained edge node devices called DeepThings. Their framework proposes a CNN partitioning scheme called the Fused Tile Partitioning (FTP) method for dividing convolutional layers. The FTP scheme fuses convolutional layers and partitions them in a vertical fashion while also reducing communication overhead.

Mao $et. al$ \cite{mao2017mednn} proposes MoDNN which is a local distributed mobile computing system for CNN  over Wireless Local Area Network (WLAN). MoDNN introduces execution parallelism among multiple mobile devices. This system uses two partition schemes for data
delivery time between the mobile devices based CNN layers and the different mobile computing capabilities in the network.
Mao $et. al$ \cite{mao2017mednn} proposes a local distributed mobile computing system called MeDNN for CNN inference on mobile devices. meDNN uses  Greedy Two Dimensional Partition (GTDP) and Structured Model Compact Deployment (SMCD) used respectively to partition and deploy the CNN on mobile devices. This system uses compression schemes to introduce structured sparsity pruning technique to further accelerate CNN inference.
Li $et. al$ \cite{li2018edge} proposes Edgent which is a co-inference framework that enables CNN partitioning to facilitate the collaboration between edge and
a mobile device to efficiently perform CNN inference.

\vspace{0mm}
\section{Conclusion}
\label{Conclusion}
\vspace{0mm}
In this paper, the security of distributed convolutional neural networks on IoT edge devices has been explored. Five different attacks on the values of the parameters have been studied.
Some attacks have more effect on the computational-centric layer and other attacks have more effect on the memory-centric layer from a classification accuracy perspective. A novel stealthy attack implementation is discussed. All the proposed attacks have reasonably less timing overhead. This work also indicates to the distributed CNN designer that which of the NDs can belong to untrusted nodes without compromising on overall CNN security.

\vspace{0mm}

\bibliographystyle{IEEEtran}
\bibliography{references}
\end{document}